\documentclass[letterpaper, 10 pt, conference]{ieeeconf}  

\IEEEoverridecommandlockouts                              
                                                          
\usepackage{amsmath} 
\usepackage{amssymb}  
\usepackage{cite}
\usepackage{color}
\usepackage{hyperref}
\usepackage{graphicx} 
\usepackage{savesym}
\savesymbol{AND}
\usepackage[ruled,vlined]{algorithm2e}
\usepackage{algorithmic}
\let\labelindent\relax
\usepackage{enumitem}
\usepackage{tikz}
\usetikzlibrary{shapes,arrows,petri,topaths}
\usepackage{amsfonts}
\usepackage{commath}
\usepackage{caption}
\usepackage{subcaption}
\usepackage[vlined,ruled]{algorithm2e}
\usepackage[utf8]{inputenc}
\usepackage{booktabs}
\usepackage{multirow}
\usepackage{xcolor}
\usepackage{balance} 
\usepackage{siunitx}
\allowdisplaybreaks

\newtheorem{definition}{Definition}
\newtheorem{remark}{Remark}

\title{\LARGE \bf
A generalized algorithm and framework for online $3$-dimensional bin packing in an automated sorting center
}

\author{Ankush Ojha, Marichi Agarwal, Aniruddha Singhal, Chayan Sarkar, Supratim Ghosh, and Rajesh Sinha
\thanks{The authors are with the Tata Consultancy Services -- Research, India. {\tt\small ojha.ankush@tcs.com, marichi.agarwal@tcs.com,  aniruddha.singhal@tcs.com, sarkar.chayan@tcs .com, supratim.ghosh2@tcs.com, rajesh.sinha@tcs.com}}}

\begin{document}
\maketitle
\begin{abstract}
Online $3$-dimensional bin packing problem (O$3$D-BPP) is getting renewed prominence due to the industrial automation brought by Industry 4.0. However, due to limited attention in the past and its challenging nature, a good approximate algorithm is in scarcity as compared to 1D or 2D problems. This paper considers real-time O$3$D-BPP of cuboidal boxes with partial information (look-ahead) in an automated robotic sorting center. We present two rolling-horizon mixed-integer linear programming (MILP) cum-heuristic based algorithms: \emph{MPack} (for bench-marking) and \emph{MPackLite} (for real-time deployment). Additionally, we present a framework \emph{OPack} that adapts and improves the performance of BP heuristics by utilizing information in an online setting with a look-ahead. We then perform a comparative analysis of BP heuristics (with and without \emph{OPack}), \emph{MPack}, and \emph{MPackLite} on synthetic and industry provided data with increasing look-ahead. \emph{MPackLite} and the baseline heuristics perform within bounds of robot operations and thus, can be used in real-time.
\end{abstract}

\section{Introduction}
With the advent of industrial robotics, automated sorting and packaging has received a lot of attention\footnote{https://www.roboticstomorrow.com/article/2019/12/packaging-the-future/14515}. Packaging problems arise in various industries such as logistics, retail, e-commerce, postal service, etc. One of the main tasks in an automated sorting center involves packing an incoming stream of boxes into large containers or bins in real-time with a high packing efficiency. This task is an evolved version of the online $3$-dimensional bin packing problem ($3$D-BPP) where a robot/robotic arm has to make decisions within a time constraint to pick up a box from the conveyor and place it in its final position within a bin. These bins are then closed and sent to various destinations.

The problem of packing boxes optimally into a large container a.k.a. bin-packing is well known to be NP-hard even in one dimension in the offline setting \cite{mG:1979}. In the offline $3$-D setting, \emph{a priori} information about the dimensions of all the boxes to be packed is available. A number of fast and nearly optimal heuristics are available for offline bin packing such as best-fit, first-fit, floor/column building, etc.~\cite{xZjB:2016,bMaA:2017,sEfG:2019}, including some robotic packing systems~\cite{sMdP:2007,dBjK:2003}. 

On the other hand, in the online setting, the packer has information about only the next upcoming box and thus, has to work with this partial information to obtain tight and efficient packing arrangement, therefore, making the problem much harder. This problem has received attention in the past (surveys may be found in \cite{hCaK:2017,mB:2014,lEmL:2010}). However, O$3$D-BPP with partial information in an industrial setting with robot-packable constraints still remains very challenging \cite{rSwT:2019}. A further problem ensues when the boxes arriving are of mixed sizes with no underlying pattern.

The problem of limited information in the online setting is alleviated to some extent if a ``look-ahead'' is available; i.e., partial information may be available about the boxes in the immediate vicinity of the packer~\cite{eG:1995}. The packing algorithm's job is to utilize this \emph{extra} information and make complex decisions so that its efficiency can be improved. Existing heuristics for O$3$D-BPP lack this capability and thus, need to be adapted to the settings with look-ahead. According to our client partners, the space utilization of bins is the most important metric for defining the success of an automation algorithm. An automated packer does not have the flexibility of reshuffling that humans have, and hence, is expected to perform worse. This results in more number of bins used; therefore, an increase in cost. 

In this article, our focus is to generate real-time O$3$D-BPP algorithms for a robotic packing agent that has the following features -- a) they utilize the (partial) information about the boxes in the look-ahead to make a better decision; b) respect robotic packing constraints on stability, orientation, and lack of reshuffling capacity; c) computationally fast enough so that the entire decision-making process is finished by the time robot is ready to pick-up a new box; and d) perform consistently well on bins and boxes of various sizes. 

The main contributions of this article are three-fold. Firstly, we develop rolling-horizon MILP-cum-heuristic based approaches \emph{MPack} (for bench-marking) and \emph{MPackLite} (real-time deployment). While \emph{MPack} produces the best packing efficiency its time-complexity is exponential compared to linear for \emph{MPackLite} (with marginal loss in efficiency as compared to \emph{MPack)}. 
Second, we develop a novel framework \emph{OPack} using \emph{virtual packing} that can adapt and improve the performance of bin packing heuristics (both offline and online) by utilizing extra information in an online setting with a look-ahead.
Finally, we compare \emph{MPack}, \emph{MPackLite}, and baseline heuristics \cite{tCgP:2008} via numerical experiments on industrial and synthetic data-sets to analyze the efficacy of the algorithms for various box and bin sizes.

\section{Description}
\label{sec:desc}

\begin{figure}[t]
    \centering
         \includegraphics[scale=0.18]{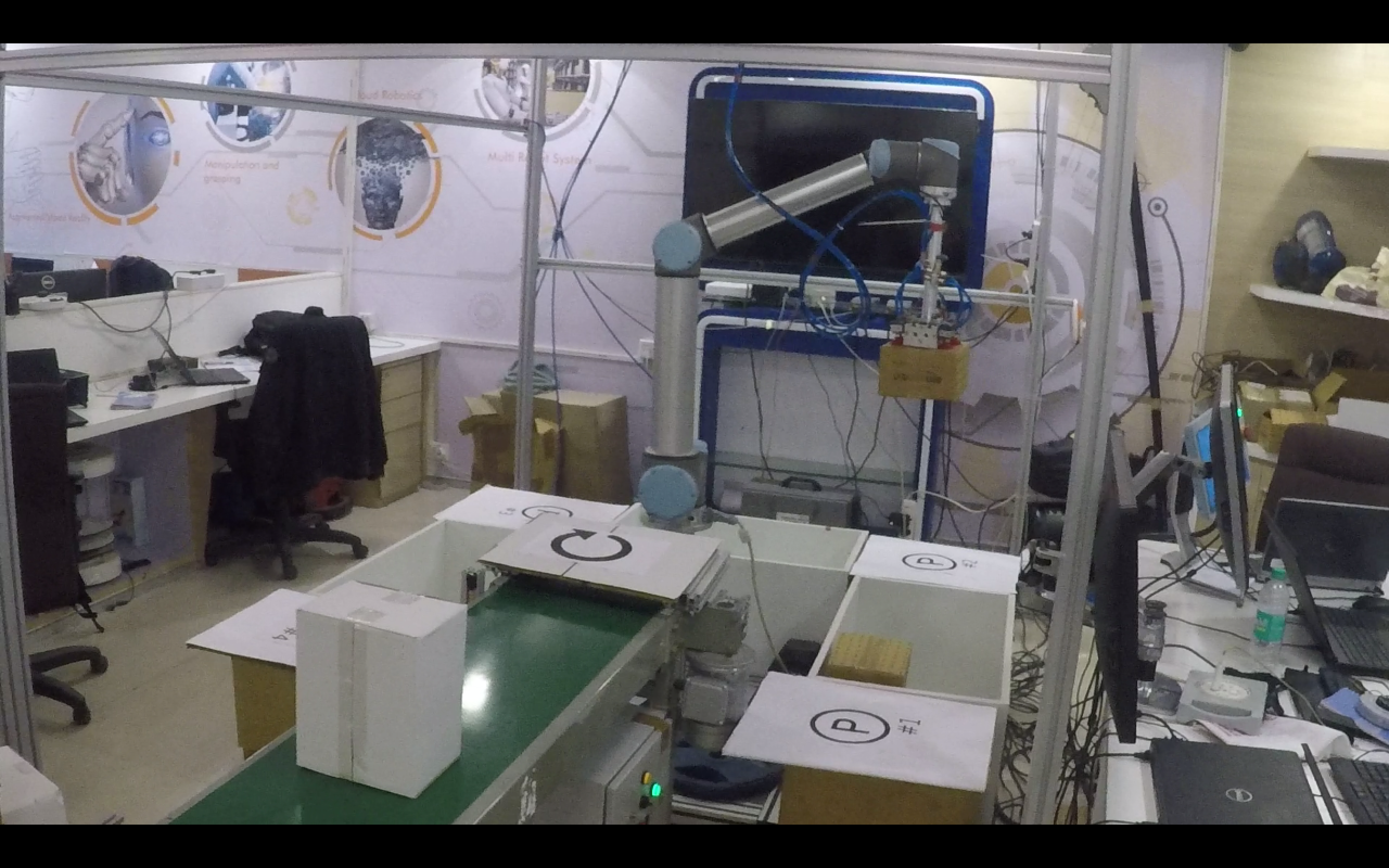}
         \caption{Lab mock-up of robotic arm packing boxes into bins.}
         \label{fig:packing_arm}
     \label{fig:system}
\end{figure}

\subsection{Sorting Center Structure}
In an automated packaging terminal, first, the arriving boxes are scanned to capture their dimensions. Very large boxes (colloquially known as ``uglies'') are usually handled manually and/or by multiple robots and are beyond the scope of this paper. We are interested in the online packaging of small/medium boxes (now onward boxes) in large bins. Separating boxes based on dimensions has the following advantages. Firstly, it helps keep the number of bins low by avoiding the situation of an ugly arriving within a stream of small-medium boxes; secondly, the same robotic arm may not be suitable for picking boxes of highly varying dimensions. After the scanning process, the boxes are placed onto the ``pick conveyor'' for placement into the open bins by a robotic arm (placed at the end of the conveyor). Fig.~\ref{fig:packing_arm} illustrates a mock-up model created in our laboratory. 
The bins can either be long-distance containers (LDCs), pallets, or roller-cages (RCs) depending upon their dimensions and use-cases. There is always a fixed number of open bins at a given time due to space constraints. Hence, before opening a new bin, one of the already open ones needs to be closed.

\subsection{Robotic Packing Framework}
The number of boxes available to the robot in look ahead is fixed depending on the pick conveyor's length, maximum box dimension to be packed, and inter-box gap. Our client partner uses a $9$ meter pick conveyor where a look-ahead of up to $6$ boxes is possible. The packing algorithm decides which box to pick considering the available boxes on the pick conveyor and the current state of the bin. The robotic arm packs the box accordingly before a new box arrives (typically $5-6$ seconds). We make the following standard assumptions. 

\begin{enumerate}
\label{ass:standard}
    \item The conveyor belt is circular or oval so that the boxes circulate on it until they are placed.
    \item The boxes are cuboidal and non-deformable.
    \item Whenever a box is picked up, one more box gets added to the conveyor so that the look-ahead remains the same.
    \item $\operatorname{Volume}$ (single box) $<=$ $\operatorname{Volume}$ (single bin).
\end{enumerate}
Additionally, there are some \textbf{robotic packing constraints}: 
\begin{itemize}[leftmargin=*]
    \item 
    The number of orientations available for placing a box is limited by the DOF of the robotic arm. The following constraint is imposed in our lab -- the box's largest dimension cannot be aligned along the arm's vertical axis.
    \item
    To ensure stability and tight packing, a box is placed on top of another (inside the bin), if at least $3$ of its vertices are supported by either other boxes (already placed) or the bin floor.
    \item
    Once a box is placed inside a container, it cannot be reshuffled to a new position (imposed to ensure real-time and quick decision making to ensure unclogged chutes).
    \item
    The robotic arm has a sensitivity of $1$cm, and thus, any decimal dimensions are rounded off to the next cm.
\end{itemize}

\subsection{Mathematical Preliminaries}
Let $L,B,H$ and $l_i,b_i,h_i$ denote the length, breadth, and height of a bin and the $i$th box, respectively. The bins are chosen to be of uniform size. To impose the packability assumption, we have $\max\{l_i, b_i, h_i\} \leq \min\{L, B, H\}$ for every $i$. The  front-left-bottom corner of the bin is treated as the origin and the interior of the bin is thus, treated as $\mathbb{R}^3$. For each box $i$ placed inside a bin, the coordinates $(x_i,y_i,z_i)$ and $(\bar{x}_i,\bar{y}_i,\bar{z}_i)$ denote the front-left-bottom and the back-right-top corners, respectively. These two points uniquely determine the position of a box inside a bin. 

The objective of the O$3$D-BPP algorithm is to minimize $N$ where $N$ denotes the number of bins used for packing the boxes. However, since there is a steady online stream of boxes arriving on a given day, it is very difficult to estimate the number of bins required for optimal packing of these boxes \emph{a priori}. Therefore, we take space utilization of the packed bins as an alternate measure. Formally, it is:

\begin{definition}
The fill-rate efficiency of an algorithm for a closed bin is the ratio of total volume of all the boxes placed inside the bin to the volume of the bin. In other words,
\begin{equation}\label{eq:fillrate}
    \text{fill-rate}=\frac{\text{total volume of boxes inside a bin}}{\text{volume of bin}}\times 100 (\%).
\end{equation}
\end{definition}
For an experiment, we describe the overall fill-rate efficiency of the algorithm for only closed bins, i.e., if $N_C$ bins are closed during the experiment, the overall efficiency of the experiment equals the average of the fill-rates of these $N_C$ bins. The fill-rate efficiency also acts as an indirect measure for \emph{asymptotic competitive (approximation) ratio} (ACR), which is the metric typically used for evaluating bin-packing algorithms. ACR is a direct measure of how many bins are used to pack the boxes in a data-set \cite{hCaK:2017}.

\section{Algorithms for $3D-BPP$}\label{sec:algo}
In this section, we describe the techniques for O$3$D-BPP using a robotic arm: a multi-objective MILP-cum-heuristic based algorithm \emph{MPack} and its derivative \emph{MPackLite}; and a framework \emph{OPack} to adapt baseline heuristics into online setting with look-ahead. Due to the robotic arm's sensitivity, the interior of the bin is converted into a $3$-D grid where every cell $C_i$ with $i\in\{1,\dots,LBH\}$ is a cube of $1$cm. 

\subsection{Multi-objective Optimization based O$3D-BPP$: MPack}
 Our first approach \emph{MPack} is based on multi-objective optimization and combines the best features of several heuristics within the ambit of MILP to produce superior solutions. 

\textbf{Objective function:} The objective function tries to capture the essence of the standard heuristics as, 
\begin{equation}\label{eq:obj}
    \min\left\{\sum_{i\in P_L}\left( w_1 (x_i+y_i) + w_2\cdot \bar{z}_i + w_3\sum_{j=1}^{N_B} p(i,j)\cdot j\right)\right\},
\end{equation}
where $P_L$ denotes the set of boxes on the conveyor serving as the look-ahead; $\ell=|P_L|$; $p(i,j)\in\{0,1\}$ is such that $p(i,j)=1$ if and only if, box $i$ is placed in bin $j$, and finally $N_B$ denotes the total number of currently open bins. Each currently open bin is given an index $j\in\{1,\dots,N_B\}$ where a smaller index implies that the bin was opened earlier.

The objective function comprises of three distinct components. Minimizing $x_i$ and $y_i$ for box $i$ amounts to minimizing the spread of the boxes on the floor (column-building). Similarly, minimizing $\bar{z}_i$ aims to reduce the overall height of packing arrangement (floor-building). The third component mimics \emph{first fit} where earlier opened bins are preferred. 

\begin{remark}\label{rem:weights}
The first and second components in the objective function act as countermeasures to each other. The non-negative weights $w_1$ and $w_2$ denote which component is given more importance and these seem to depend on the ratio of the bin's height to its base area. For instance, if the bins in use have large height to surface area ratio; then $w_1$ should be greater than $w_2$, and vice versa. An analytic way to determine the optimal choice of weights corresponding to a height to surface area ratio is currently under consideration. On the other hand, $w_3$ is chosen very high (between $10$ and $100$) since the usual values of $x_i,y_i,\bar{z_i}$ in cms are $2$ orders of magnitude higher than $p(i,j)$.
\end{remark}

\subsubsection{Constraints}\label{sec:constraints}

The algorithm \emph{MPack} converts O$3$D-BPP with look-ahead into a sequence of offline bin packing problems. Assume that at time $t$ with $|P_L|=\ell$, $N_B$ bins are currently open, and $N$ boxes already placed in the open bins. The offline bin packing problem is to minimize the objective function~\eqref{eq:obj} subject to packing constraints (described later) for the $N+\ell$ boxes to be placed in $N_B$ bins. However, since the packed boxes can not be reshuffled, their corresponding variables are fixed to values from the previous run. For instance, at $t=0$, the objective is to pick and place one of the first $\ell$ boxes in the first $N_B$ bins. After the box is placed, at $t=1$, one more box gets added to the conveyor. Thus, at $t=1$, the optimization problem consists of $\ell + 1$ (look-ahead and one already packed) boxes to be placed into $N_B$ open bins. The variables corresponding to the box already packed are fixed to the values obtained at $t=0$. This procedure is repeated iteratively until all the boxes are packed. Note that since $\ell$ and $N_B$ are fixed, the size of the problem is bounded.

The constraints for \emph{MPack} can be divided into three sets -- geometric constraints, vertical stability constraints~\cite{cPmS:2016}, and efficient packing constraints. To maintain the brevity of the paper, we will present only a brief sketch of the most important geometric and vertical stability constraints for the sake of completeness (for details please refer to~\cite{cPmS:2016}). For a single iteration of \emph{MPack}, the parameters $\ell$, $N$, and $N_B$ are defined previously. Let $P_B=\{1,\dots,N\}$ denote the set of boxes already placed, and $P_L=\{N+1,\dots,N+\ell\}$ denote the set of boxes in the look-ahead, respectively. Similarly, the sets $P\triangleq P_B\cup P_L$ and $O_B=\{1,\dots,N_B\}$ denotes the set of boxes and bins used in the current iteration. Every time a new bin is opened, an old bin is closed and the bin indices are reset. For all $i,k\in P$; $j\in\{1,\dots,N_B\}$; and $s\in\{x,y,z\}$ (coordinates), let 
\begin{align*}
p_{ij} &= \left\{
                \begin{array}{l}
                1 \text{ if box \emph{i} is placed in container \emph{j}},\\
                0 \text{ otherwise,}
                \end{array}
            \right. \\
u_{j} &= \left\{
                \begin{array}{l}
                1 \text{ if container \emph{j} is used},\\
                0 \text{ otherwise,}
                \end{array}
            \right. \\
s_{il}^p  &=  \left\{ \begin{array}{l}
         1 \text{ if box $i$ is on the right of box $k$} \ \left( \bar{s}_k \leq s_i \right), \\
         0 \text{ otherwise} \left( s_i < \bar{s}_k\right).
    \end{array}
    \right.
\end{align*}

For each $i\in P$ and for every $a,b\in\{x,y,z\}$, we define
\[
r_{iab}  = \left\{ \begin{array}{l}
         1 \text{ if the side \emph{b} of box \emph{i} is along \emph{a}-axis of the bin}, \\
         0 \text{ otherwise.}
    \end{array}
    \right.
\]
The variable, $r_{iab}$ checks the degree of freedom of the robotic arm, which is used to place the box in the containers. For example, if the length of the box is kept along the length of the container then, $r_{ixx}$ = 1.

\textbf{Geometric constraints.} These ensure a box is always placed within a bin and all feasible orientations are respected:
\begin{align}\label{eq:gc1}
\sum_{i\in P_B\cup P_L} p_{ij} \leq & u_j,\quad \sum_{j=1}^B p_{ij} = 1, \quad s_i' \leq \sum_{j=1}^B L_j \, p_{ij}, \nonumber\\
\bar{x}_i - x_i &= r_{ixx} \, l_i + r_{ixy} \, b_i + r_{ixz} \, h_i, \nonumber\\
\bar{y}_i - y_i &= r_{iyx} \, l_i + r_{iyy} \, b_i + r_{iyz} \, h_i, \nonumber\\
\bar{z}_i - z_i &= r_{izx} \, l_i + r_{izy} \, b_i + r_{izz} \, h_i, \nonumber\\
\sum_{b} r_{iab} &= 1, \quad \sum_{a} r_{iab} = 1,
\end{align}
for all $i\in P_B\cup P_L$, $j\in O_B$, $s, a, b\in\{x,y,z\}$. 

The $2^{nd}$ to $4^{th}$ lines in \eqref{eq:gc1} describe the possible orientations of a box before placing it. In case restrictions are to be imposed, corresponding $r$ should be set equal to $0$. 
\textbf{Vertical stability constraints.}
These constraints ensure that we have a stable packing arrangement such that the box being placed gets adequate support at the bottom and does not float in the air. This is ensured by placing a box either on the floor of the container or in a location where at least 3 vertices of the base of the box are \emph{supported} by underlying boxes. Please refer to \cite{cPmS:2016} for mathematical details.

\textbf{Constraints for efficient packing} (CEP).
These constraints ensure that there are no gaps (or ``holes'') in the packing arrangement, i.e., whenever a new box is placed, two of its surfaces should either be touching the already placed boxes (along the X-Y plane) or the walls of the bin.  

The following variables are used for these constraints:

\begin{align*}
d_{ikc} & =  \left\{ \begin{array}{l}
         \text{1 if box $i$ is in contact with box $k$ along $c$ axis,} \\
         \text{0 otherwise,}
    \end{array}
    \right.\\
d_{ijw}^w & =  \left\{ \begin{array}{l}
         \text{1 if box $i$ is in contact with wall $e$ of container $j$,} \\
         \text{0 otherwise,}
    \end{array}
    \right.
\end{align*}
for all $i\in P_L$, $k \in P$, $j \in O_B$, $c \in \{x,y\}$ and $w \in \{1,...,4\}$. Restricting our attention to the X-Y plane, we can see that there are $4$ possible walls of the bin with which a box can make contact and hence let $w\in\{1,\dots,4\}$. The constraints are formulated as:
\begin{align}\label{eq:eff1}
c_i \leq \bar{c}_k + (1 - d_{ikc}) \, D,&\quad
c_i \leq \bar{c}_k + (d_{ikc} - 1) \, D,\nonumber\\
c_i \leq (m - 1)\,D &+ (1 - d_{ijp}^w) \, D,\nonumber\\
c_i \leq (m - 1)\,D &+ (d_{ijp}^w - 1) \, D,\nonumber \\
\bar{c}_i \leq m D + (1-d^w_{ijq})D,&\quad
\bar{c}_i \leq m D + (d^w_{ijq}-1)D,
\end{align}
for every $i\in P_L$, $k\in P$, $j\in O_B$, $c\in\{x,y\}$. For $c=x$; $D=L$, $m=j$, $p=1$ and $q=3$ whereas, for $c=y$; $D=B$, $m=1$, $p=3$, and $q=4$, respectively. In addition, we need the following set of constraints:
\begin{align}\label{eq:eff2}
\sum_{i\in P_L} \sum_{j = 1}^{N_B} (d_{ij1}^w + d_{ij3}^w) + \sum_{i\in P_L} \sum_{k\in P_B\cup P_L} d_{ikx} &\leq 1,\nonumber\\
\sum_{i\in P_L} \sum_{j = 1}^{N_B} (d_{ij2}^w + d_{ij4}^w) + \sum_{i\in P_L} \sum_{k\in P_B\cup P_L} d_{iky} &\leq 1,\nonumber\\
\sum_{i\in P_L} \sum_{j = 1}^{N_B} \sum_{e = 1}^4 d_{ije}^w + \sum_{i\in P_L}^{\ell} \sum_{k\in P_B\cup P_L} \sum_{c\in\{x,y\}}  d_{ikc} &\geq 2.
\end{align}
In addition to ensuring CEP, there is also a need to remove redundancies within the formulation which may arise from counting the same surface twice if it touches two or more boxes. This is enforced by \eqref{eq:eff2}.

\subsubsection{Algorithm}
For a box $B$, let $Loc_B$ define the location coordinates $(x_B,y_B,z_B)$ calculated by following the packing algorithm steps. For a set $\mathcal{B}\triangleq \{B_1,\dots,B_n\}$, $Loc_{\mathcal{B}}$ denotes the collection $\{Loc_{B_1},\dots,Loc_{B_n}\}$. Briefly, the steps of \emph{MPack} are following. 
\begin{itemize}[leftmargin=*]
    \item First, try to find $Loc_{P_L}$ by minimizing \eqref{eq:obj} subject to \eqref{eq:gc1}--\eqref{eq:eff2}. If $Loc_{P_L}$ is found, choose one box from $P_L$ using a selection heuristic (described later), pick it and place it.
    \item If $Loc_B$ is not found for any $B\in P_L$, then repeat the above procedure by removing the box farthest from the arm from $P_L$ (i.e., decreasing $\ell$ by one) until $Loc_{P_L}$ is found or $|P_L|=1$.
    \item If $|P_L|=1$, evaluate each box on the conveyor individually to find feasible locations. If any of them are found, go for box selection and placement. Else, close the bin with the highest fill-rate, open a new bin and repeat steps $1$--$3$.
\end{itemize}

The algorithm terminates every time provided there are enough bins. If there is no feasible location in the existing bins, a new bin is opened (after closing an earlier bin) and this ensures that the feasible location for at least one of the boxes is always found (in the worst case, it's in the new bin).

\subsubsection{Box Selection Heuristic}\label{sec:boxpack}

Whenever $Loc_B$ for multiple $B\in P_L$ are found, the robotic arm has to choose one among them. One way to do this is to select the box that yields minimal increase in $w_1(x_i+y_i)+w_2\cdot\bar{z}_i$ (to align with the objective) and can be supported solely by the boxes in $P_B$ and bin (does not require support from $P_L$) to ensure stability\footnote{Another heuristic may involve choosing a large box when bin fill-rate is low and a small box when it is high.}.

\subsection{OPack: Framework for O$3$D-BPP}

\setlength{\textfloatsep}{0pt}
\begin{algorithm}[t]
    \caption{($B_i$,$Loc_{B_i}$) = OPack(ALGO($P_B$, $P_L$, $Loc_{P_B}$))}
    \label{algo:OPack}
    \begin{algorithmic}
    \STATE Set $\ell=|P_L|$ and $count=1$.
    \STATE Set $\tilde{P}_B=P_B$ and $Loc_{\tilde{P}_B}=Loc_{P_B}$.
    \STATE $\tilde{P}_L\triangleq \{B_1,\dots,B_\ell\}=\operatorname{SORT}(P_L)$.
    \WHILE{$count<=\ell$}
        \STATE $Loc_{B_{count}}=\operatorname{PACKRULE}(\tilde{P}_B,Loc_{\tilde{P}_B},B_{count})$.
        \STATE Update $\tilde{P}_B=\tilde{P}_B\cup\{B_{count}\}$. 
        \STATE Update $Loc_{\tilde{P}_B}=Loc_{P_{B}}\cup Loc_{B_{count}}$.
        \STATE $count$ = $count+1$.
    \ENDWHILE
    \STATE Choose $B_i\in\tilde{P}_L$ using $\operatorname{BOXPACK}$ heuristic.
    \STATE Pick $B_i$, place it at $Loc_{B_i}$, and update $P_B$, $Loc_{P_{B}}$, $P_L$.
\end{algorithmic}
\end{algorithm}

We now present a framework \emph{OPack} that can adapt standard bin packing algorithms (both offline and online) to the O$3$D-BPP with look-ahead using the concept of \emph{virtual packing}. A sketch of \emph{OPack} is provided in algorithm \ref{algo:OPack} and the following sub-routines are needed for its execution:

\begin{itemize}[leftmargin=*]
    \item $\operatorname{SORT}$: Create a \emph{virtual} sorted list of boxes in in $P_L$ according to a pre-specified rule; for e..g, decreasing order of volume or area (ties broken by decreasing height) \cite{tCgP:2008}.
    \item $\operatorname{PACKRULE}$: Calculate $Loc_B$ using $\operatorname{ALGO}$'s packing rule. For e.g., prefer earliest opened bin in First-fit (FF) or bin with maximum fill-rate in best-fit (BF).
    \item $\operatorname{BOXPACK}$: Pick a box $B$ for final packing based on: (i) stand-alone stability: $Loc_B$ should be supported by boxes in $P_B$ and the bin (not requiring any support from $P_L$); and (ii) box-selection rule in case of multiple stand-alone stable boxes.
\end{itemize}

\emph{OPack} essentially works by converting a problem of $\ell$ look-ahead into a sequence of $\ell$ problems of look-ahead $1$. However, the solution of each problem in the sequence is dependent on the former. This is achieved by a virtual update of bin state after every iteration of $\operatorname{PACKRULE}$ by placing the box under consideration in the current iteration at its feasible location. Note that this step is \emph{virtual}; i.e., this does not happen in reality on the conveyor. This \emph{virtual packing} step enables the algorithm to utilize the information about the boxes in $P_L$. Standard bin packing heuristics like First-fit (FF) or Best-fit (BF) lack this step and hence are not able to utilize the \emph{extra} information from $P_L$. We will empirically show that the use of \emph{virtual packing} in \emph{OPack} results in much better performance for algorithms than without it.

We apply \emph{OPack} to \emph{MPack}, first-fit, best-fit~(using extreme points \cite{tCgP:2008}), and Jampack (using corner points~\cite{mAsB:2020}) to yield \emph{MPackLite}, O-FF, O-BF, and O-JP, respectively. In should be noted that while FF and BF are online heuristics, they are not equipped to deal with additional information in the form of look-ahead. On the other hand, first-fit decreasing (FFD) and best-fit decreasing (BFD) use pre-sorted list of boxes which arrive one by one; and thus, are not online heuristics~\cite{jBjB:2018}. 

\subsection{MPackLite: Applying OPack to MPack}

By applying the framework \emph{OPack} to algorithm \emph{MPack}, we obtain a real-time deployable algorithm: \emph{MPackLite}. Mainly, the three subroutines in \emph{MPackLite} are:
\begin{itemize}[leftmargin=*]
    \item $\operatorname{SORT}:$ Sort boxes in $P_L$ virtually by decreasing volumes with ties broken by decreasing heights.
    \item $\operatorname{PACKRULE}:$ Evaluate $Loc_{B_{count}}$ for $B_{count}\in\tilde{P}_L$ by minimizing \eqref{eq:obj} subject to constraints in section \ref{sec:constraints}. The input parameters are $\tilde{P}_B$ and $Loc_{\tilde{P}_B}$.
    \item $\operatorname{BOXPACK}:$ Select and pack one box in $\tilde{P}_L$ satisfying stand-alone stability and selection rule (section \ref{sec:boxpack}).
 \end{itemize}
Since \emph{MPackLite} decomposes a problem of look-ahead $\ell$ to a sequence of $\ell$ problems with look-ahead of size $1$ (with dependencies between them), the computation time/box grows linearly for \emph{MPackLite} compared to exponential for \emph{MPack}. To speed up \emph{MPackLite} further, we drop the constraints on efficient packing (CEP), since they do not offer any distinct advantage for a problem with look-ahead of size $1$. The objective function (which captures floor and column building) and the constraints are enough to ensure that the packing arrangement is tight without any gaps.


\section{Experiments and Results}

\begin{table*}
    \footnotesize
  \caption{Performance comparison for: First-fit (FF), Best-fit (BF) \emph{OPack}-FF (O-FF), \emph{OPack}-BF(O-BF), \emph{OPack}-Jampack (O-JP), \emph{MPackLite} (MPL), \emph{MPack} (MP). FR($\%$) = mean fill-rate $\%$; T(s) = time/box (sec) ( ``\textemdash" indicates data not available).} 
  \label{tab:output}
  \centering
    \footnotesize{
  \begin{tabular}{cc rrrrrrrrrrrrrr}\toprule

      \textit{$\ell$} & \textit{Bin}& \multicolumn{2}{c}{\textit{FF}} & \multicolumn{2}{c}{\textit{O-FF}}& \multicolumn{2}{c}{\textit{BF}} &  \multicolumn{2}{c}{\textit{O-BF}}&  \multicolumn{2}{c}{\textit{O-JP}} & \multicolumn{2}{c}{\textit{MPL}} & \multicolumn{2}{c}{\textit{MP}}\\ 
        & & \textit{FR(\%)} & \textit{T}(s) &\textit{FR(\%)} & \textit{T}(s) &\textit{FR(\%)} & \textit{T}(s) &\textit{FR(\%)} & \textit{T}(s) &\textit{FR(\%)} & \textit{T}(s) &\textit{FR(\%)} & \textit{T}(s) &\textit{FR(\%)} & \textit{T}(s) \\ \midrule
      \multirow{ 5}{*}{1}& LDC & 64.91 & 0.045 & 64.91 & 0.044 & 62.71 & 0.05 & 62.71 & 0.05 & 60.77 & 0.023 & 68.01 & 0.359 & 68.26 & 0.62 \\
      & RC & 49.51 & 0.057 & 49.51 & 0.057 & 43.47 & 0.062 & 43.47 & 0.064 & 31.71 & 0.039 & 57.45 & 0.67 & 57.57 & 0.9\\
      & PAL & 67.43 & 0.099 & 67.43 & 0.102 & 65.17 & 0.113 & 65.17 & 0.115 & 69.13 & 0.053 & 72.89 & 0.55 & 73.38 & 1.21\\
      & EQ & 59.67 & 0.036 & 59.67 & 0.035 & 59.04 & 0.038 & 59.04 & 0.4 & 49.77 & 0.017 & 67.06 & 0.35 & 65.3 & 0.88\\
      & SYN & {\textemdash} & {\textemdash} & 68.71 & 0.049 & {\textemdash} & {\textemdash}  & 65.45 & 0.062 & 64.57 & 0.031 & 73.11 & 1.045 & 73.92 & 0.35\\ \hline
      \multirow{ 5}{*}{2}& LDC & 64.11 & 0.045 & 65.67 & 0.044 & 63.31 & 0.05 & 64.21 & 0.054 & 62.11 & 0.02 & 68.38 & 0.71 & 69.99 & 1.29 \\
      & RC & 50 & 0.056 & 50.17 & 0.053 & 43.57 & 0.063 & 44.5 & 0.07 & 32.42 & 0.032 & 58.1 & 1.29 & 59.91 & 4.02\\
      & PAL & 67.99 & 0.097 & 68.27 & 0.090 & 64.86 & 0.113 & 66.05 & 0.119 & 69.48 & 0.046 & 73.25 & 1.06 & 73.88 & 2.93\\
      & EQ & 60.33 & 0.036 & 59.75 & 0.039 & 57.74 & 0.038 & 60.01 & 0.47 & 51.68 & 0.016 & 66.39 & 0.68 & 68.54 & 1.45\\
      & SYN & {\textemdash} & {\textemdash} & 69.99 & 0.052 & {\textemdash} & {\textemdash}  & 66.49 & 0.073 & 65.22 & 0.028 & 74.05 & 2.04 & 75.03 & 0.615\\ \hline
      \multirow{ 5}{*}{3}& LDC & 65.94 & 0.045 & 67.42 & 0.05 & 63.55 & 0.05 & 64.71 & 0.067 & 61.84 & 0.021 & 69.06 & 1.159 & 70.91 & 3.22 \\
      & RC & 50.15 & 0.055 & 51.67 & 0.066 & 43.7 & 0.062 & 46.19 & 0.089 & 34.37 & 0.033 & 57.94 & 1.93 & 61.33 & 17.14\\
      & PAL & 67.33 & 0.097 & 68.83 & 0.099 & 65.62 & 0.113 & 66.76 & 0.139 & 69.12 & 0.047 & 73.6 & 1.53 & 73.53 & 9.73\\
      & EQ & 60.55 & 0.036 & 62.6 & 0.048 & 58.67 & 0.038 & 60.96 & 0.59 & 51.24 & 0.017 & 67.56 & 1.03 & 69.74 & 3.08\\
      & SYN & {\textemdash} & {\textemdash} & 70.44 & 0.06 & {\textemdash} & {\textemdash}  & 67.57 & 0.09 & 65.03 & 0.029 & 74.83 & 3.075 & 75.95 & 1.33\\ \hline      
      \multirow{ 5}{*}{4}& LDC & 66.21 & 0.045 & 68.06 & 0.063 & 63.34 & 0.05 & 67.11 & 0.082 & 61.57 & 0.022 & 70.12 & 1.507 & 71.29 & 10.71 \\
      & RC & 50.3 & 0.057 & 51.7 & 0.083 & 44.45 & 0.062 & 47.6 & 0.109 & 34.77 & 0.036 & 59.38 & 2.54 & 62.52 & 58.92\\
      & PAL & 68.26 & 0.097 & 69.67 & 0.115 & 64.92 & 0.113 & 67.37 & 0.167 & 68.81 & 0.052 & 74.19 & 2.54 & 74.44 & 31.24\\
      & EQ & 59.58 & 0.036 & 61.59 & 0.057 & 59.01 & 0.038 & 61.4 & 0.7 & 50.16 & 0.018 & 67.84 & 1.4 & 70.77 & 7.19\\
      & SYN & {\textemdash} & {\textemdash} & 71.28 & 0.075 & {\textemdash} & {\textemdash}  & 68.49 & 0.112 & 64.77 & 0.03 & 75.4 & 4.16 & 76.96 & 3.35\\ \hline       
      \multirow{ 5}{*}{5}& LDC & 65.89 & 0.045 & 69.18 & 0.075 & 63.48 & 0.05 & 66.67 & 0.097 & 61.38 & 0.025 & 70.2 & 1.877 & 72.41 & 52.55 \\
      & RC & 51 & 0.057 & 52.89 & 0.1 & 44.26 & 0.062 & 48.57 & 0.131 & 34.98 & 0.038 & 59.63 & 3.21 & {\textemdash} & {\textemdash}\\
      & PAL & 68.44 & 0.097 & 70.24 & 0.13 & 65.16 & 0.113 & 67.23 & 0.191 & 68.71 & 0.057 & 74.5 & 2.9 & 74.34 & 133.56\\
      & EQ & 61.17 & 0.036 & 62.85 & 0.065 & 59.12 & 0.038 & 62.55 & 0.084 & 51.64 & 0.019 & 69.18 & 1.78 & 71.73 & 15.38\\
      & SYN & {\textemdash} & {\textemdash} & 71.74 & 0.09 & {\textemdash} & {\textemdash}  & 68.96 & 0.137 & 64.51 & 0.033 & 75.99 & 5.06 & 77.93 & 14.64\\ 
    \bottomrule

  \end{tabular}}
\end{table*}

\begin{figure*}
    \centering
    \centering
     \begin{subfigure}[b]{0.375\textwidth}
         \centering
         \includegraphics[width=\textwidth]{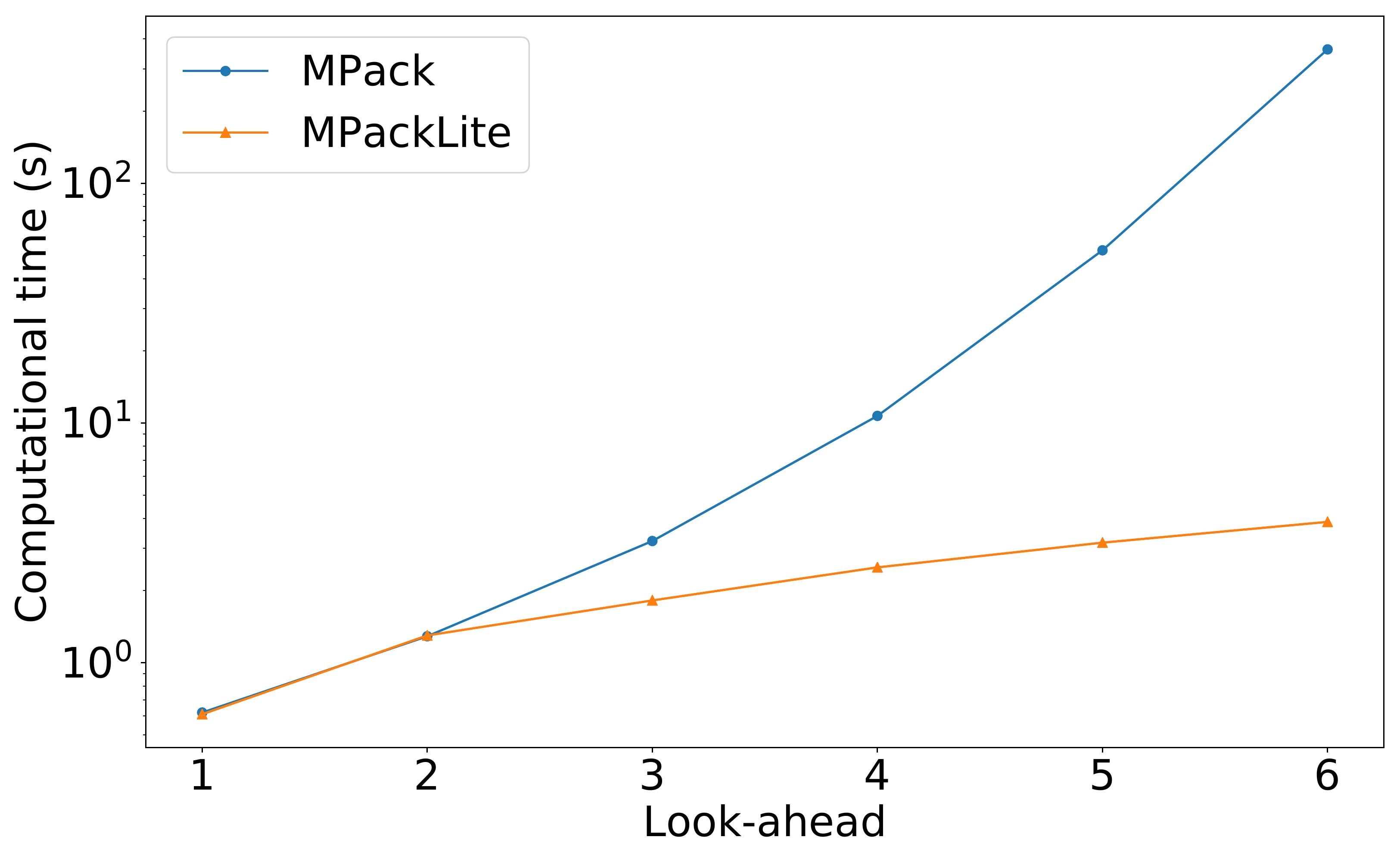}
         \caption{Comparison of computations times for LDC}
         \label{fig:comptime}
     \end{subfigure}
     \hspace{2cm}
     \begin{subfigure}[b]{0.375\textwidth}
         \centering
         \includegraphics[width=\textwidth]{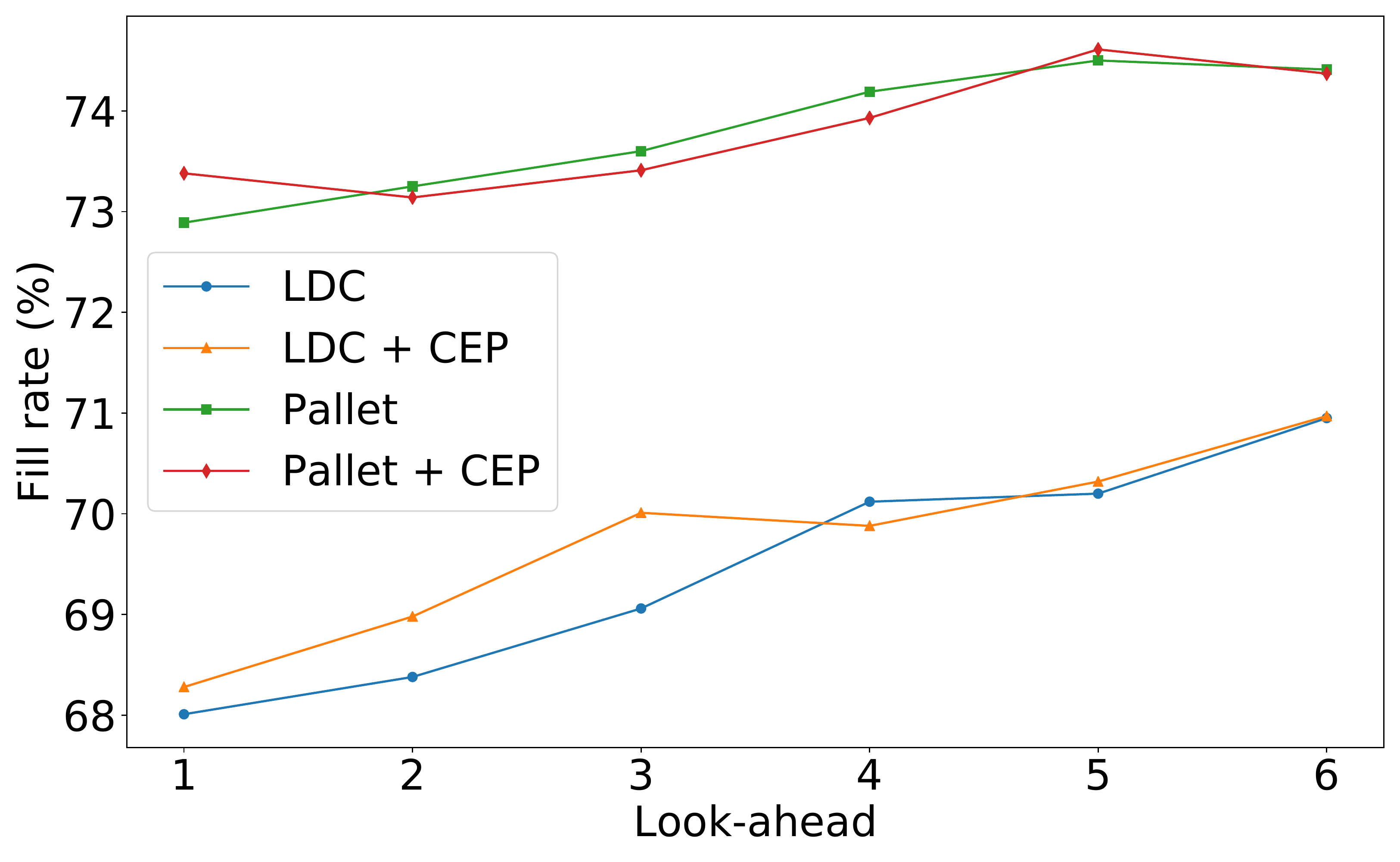}
         \caption{Effect of CEPs}
         \label{fig:cep}
     \end{subfigure}
     \caption{Comparison of various aspects of \emph{MPackLite} and \emph{MPack}.}
     \label{fig:system}
\end{figure*}

We compare the performance of \emph{MPack} (MP) and \emph{MPackLite} (MPL) against two baseline heuristics -- First-fit (FF) \& Best-fit (BF)) along with their adaptation with \emph{OPack}, i.e.,  \emph{OPack}-FF (O-FF) and \emph{OPack}-BF(O-BF). Additionally, we also compared with \emph{OPack} adaptation of Jampack (O-JP), which is a state-of-the-art offline bin packing algorithm. We performed experiments with two types of data.

\subsubsection{Industrial data} Our client partner provided data that contained dimensions of boxes from their sorting center. We generated $100$ collections of boxes by random sampling ($25$ for each industrial bin type): Long Distance Container (LDC) - $120\times 80\times 80 cm^3$, Roller-cage (RC) - $120\times 70\times 160cm^3$, Pallet (PAL) - $220\times 120\times 80cm^3$, and Equal (EQ) - $80\times80\times80 cm^3$. The total volume of each collection would fill $4$ bins. The number of boxes in collections varies according to the bin dimensions. In addition, the box dimensions are not integers and hence, a $\operatorname{ceiling}$ function is used while placing the boxes. However, the fill rate is calculated based on actual values.
\subsubsection{Synthetic data} We generated $30$ collections of boxes such that each collection would optimally fill $10$ bins with dimensions $80\times 45\times 45 cm^3$. The box dimensions are integers; thus, accurate placement can be achieved.

The boxes in each collection are sent on the conveyor in completely random order. All $6$ rotations are allowed, and at any time, up to $3$ bins are open (bin with highest fill-rate is closed in case a new bin is required to be opened). 

\textbf{Results.}
The experiments were run on an $8$-core laptop with i7-6820HQ processor (2.7 GHz speed/16GB memory). The weights $w_1$, $w_2$, and $w_3$ in Eq.~\ref{eq:obj} are set to $1$, $1$, and $100$, respectively, for \emph{MPack} and \emph{MPackLite}. The results for mean fill-rates (MFRs) and mean computation time/box are captured in Table \ref{tab:output}. Whenever experiments were stopped either due to enormous computation time or lack of ability to provide new information, their entries are marked using ``\textemdash''. 

\textbf{Discussion.}
\emph{MPack} performs the best, at the cost of increased computation time; and hence, is used for benchmarking. Among the others, \emph{MPackLite} performs the best against heuristics for all data sets and bin-sizes. For instance, it achieves a relative improvement ranging from $1.5\%$ (LDC) to $12.75\%$ (RC) against O-FF; and $5.3\%$ (LDC) to $23\%$ (RC) against O-BF for $\ell=5$ with its computation time is between $10$ and $30$ times higher. However, all the computation times for \emph{MPackLite} are within the bounds of robot operation ($5-6$ seconds/box) and hence, it can be deployed in real-time. The MFRs for synthetic data sets are higher than real data sets for all algorithms because their boxes have integral dimensions. Also, one can observe that while \emph{MPackLite} performs marginally lower than \emph{MPack} in terms of MFRs; its computation time growth is linear compared to exponential for \emph{MPack} (Fig.~\ref{fig:comptime}). Additionally, as Fig.~\ref{fig:cep} shows, CEPs do not offer any advantage to \emph{MPackLite}.

A comparison between FF/BF and O-FF/O-BF reveals the benefits of using \emph{OPack} with virtual packing. While the MFRs for FF and BF are stagnant with increasing $\ell$; the same show an increasing trend for O-FF and O-BF. For instance, for the case of LDC, the MFR of BF and O-BF vary from $62.71\%$ to $63.48\%$ and from $62.71\%$ to $66.67\%$, respectively, from $\ell=1$ to $\ell=5$ (resulting in a relative improvement of $\approx 4.8\%$ for O-BF over BF). We also observe a similar behavior across bin dimensions. This clearly shows how virtual packing helps utilize information from $P_L$. On the other hand, due to the extra steps of virtual packing, the computation time/box also increases with respect to $\ell$.

Table~\ref{tab:output} shows that O-FF performs better than O-BF and O-JP. This is due to the setup where the bins opened earlier with the highest-fill rates have to be closed for new ones to be opened. Thus, preferring the earliest opened bins (rule of FF) yields the highest efficiency. One can conjecture that if all the available bins are opened concurrently, then O-BF and O-FF would perform similarly since $ACR(FF)=ACR(BF)=1.7$.
\section{Conclusions}\label{sec:conc}
In this paper, we presented two approaches for solving the online robotic bin-packing problem in an automated sorting center with partial information about upcoming boxes: (i) an algorithm \emph{MPackLite}; (ii) a framework \emph{OPack} which can adapt baseline algorithms (online or offline) to this setup. \emph{OPack} helps BP algorithms to extract and utilize this partial information; thus, improving their performance. Future extensions of this work are twofold: achieving an even higher fill-rate by working across packaging levels, and deriving a theoretical ACR for \emph{MPack}/\emph{MPackLite}.

\bibliographystyle{IEEEtran}
\bibliography{main.bib}

\end{document}